\documentclass[acmsmall,screen,manuscript]{acmart}

\usepackage{graphicx}
\usepackage{algorithm}
\usepackage{algorithmic}
\usepackage{amsmath}
\AtBeginDocument{%
  }

\setcopyright{acmcopyright}
\copyrightyear{2024}
\acmYear{2024}
\acmDOI{XXXXXXX.XXXXXXX}

\usepackage{natbib}

\begin{document}

\title{Reducing Labeling Costs in Sentiment Analysis via Semi-Supervised Learning}

\author{Minoo Jafarlou}
\email{mjafarlou1@gsu.edu}
\orcid{0002-3916-0242}
\affiliation{%
  \institution{Georgia State University}
  \streetaddress{P.O. Box 3965}
  \city{Atlanta}
  \state{Georgia}
  \country{USA}
  \postcode{30302-3965}
}

\author{Mario M. Kubek}
\affiliation{%
  \institution{Georgia State University}
  \streetaddress{P.O. Box 3965}
  \city{Atlanta}
  \state{Georgia}
  \country{USA}}
\email{mkubek@gsu.edu}

\renewcommand{\shortauthors}{Jafarlou et al.}

\begin{abstract}
Labeling datasets is a noteworthy challenge in machine learning, both in terms of cost and time. This research, however, leverages an efficient answer. By exploring label propagation in semi-supervised learning, we can significantly reduce the number of labels required compared to traditional methods. We employ a transductive label propagation method based on the manifold assumption for text classification. Our approach utilizes a graph-based method to generate pseudo-labels for unlabeled data for text classification task, which are then used to train deep neural networks. By extending labels based on cosine proximity within a nearest neighbor graph from network embeddings, we combine unlabeled data into supervised learning, thereby reducing labeling costs. Based on previous successes in other domains, this study builds and evaluates this approach's effectiveness in sentiment analysis, presenting insights into semi-supervised learning.

\end{abstract}



\received{8 August 2024}
\received[revised]{5 October 2024}
\received[accepted]{20 September 2024}

\maketitle

\section{Motivation}
Obtaining labeled data for real-world applications is costly and time-consuming due to the need for human labor and specialized knowledge. In contrast, unlabeled data is abundant and inexpensive. This research uses label propagation, a graph-based semi-supervised learning method, to enhance performance in sentiment analysis tasks using deep learning techniques. Label propagation generates pseudo-labels for unlabeled data based on the proximity of data points within a graph, enabling cost-effective learning by leveraging data closeness and reducing the dependence on extensive labeled datasets\cite{iscen2019label}.

The study focuses on achieving high performance with limited labeled data, a crucial aspect for industries where data labeling is scarce yet accuracy is critical. By experimenting with various label propagation parameters and techniques, such as similarity metrics, graph construction, and propagation strategies, we aim to tailor performance enhancements for specific natural language processing (NLP) tasks. We also emphasize the generalizability of our approach across different NLP and natural language understanding (NLU) tasks, highlighting its strengths and scalability in handling large datasets and complex models.

The primary contribution of this paper is to demonstrate how label propagation can improve sentiment analysis by utilizing the abundance of unlabeled data, thereby reducing labeling costs. We hypothesize that label propagation will significantly enhance the accuracy of sentiment analysis models by effectively leveraging unlabeled data, resulting in performance comparable to fully supervised models trained on large labeled datasets. Label propagation has shown promising results in other areas of deep learning, and we aim to assess its efficacy specifically within the sentiment analysis NLP task.

The key metric for evaluating the effectiveness of label propagation in this project is the accuracy and F1 score. The potential benefits of semi-supervised learning (SSL) in NLP include mitigating data imbalance, adapting to continuously collected data, and leveraging recent advancements in SSL techniques like self-training, consistency regularization, and pseudo-labeling. These methods can improve the quality of pseudo-labels, making SSL more effective and reliable, and lead to the development of robust, high-performing models that generalize well to unseen data.

This paper is structured as follows: we begin with a literature review, followed by an exploration of the fundamentals and concepts of our approach. We then detail our experimental results and conclude with a summary of our findings.

\section{Literature Review}
Semi-supervised learning (SSL) is an intermediate between supervised and unsupervised learning, utilizing a small amount of labeled data alongside a large amount of unlabeled data to improve learning performance. Standard SSL methods include self-training, co-training, multi-view learning, and Transductive Support Vector Machines (TSVMs). These methods exploit the data's inherent structure to propagate labels from labeled to unlabeled instances, improving model performance with minimal labeled data \cite{reddy2018semi}. 

Graph-based methods in SSL leverage the relationships between data points, defining them as nodes in a graph. The edges between nodes are weighted based on similarity measures, such as word order, context similarity, and word frequency \cite{widmann2017graph}. Graph Neural Networks (GNNs) have been remarkably effective in this domain, performing state-of-the-art results for node classification. InfoGNN, a novel informative pseudo-labeling framework, maximizes mutual information to label the most informative nodes, improving performance with few labels \cite{li2023informative}. Another approach, as studied in \cite{iscen2019label}, utilizes a transductive label propagation method based on the manifold assumption. This method iterates between creating a nearest neighbor graph of the dataset and generating pseudo-labels based on the network embeddings, improving performance on several datasets, particularly in the few labels scenarios.

Pseudo-labeling includes using a model's predictions on unlabeled data as true labels. This technique is specifically effective in SSL, where pseudo-labels can guide the training of deep neural networks \cite{lee2013pseudo}. The method encourages low-density detachment between classes, a principle known as entropy regularization. Recent advancements, such as PARS (Pseudo-Label Aware Robust Sample Selection), combine pseudo-labeling with sample selection and noise-robust loss functions to handle noisy labels and improve model accuracy significantly \cite{goel2022pars}. \cite{tarvainen2018meanteachersbetterrole} presents Temporal Ensembling and Mean Teacher as advanced techniques in semi-supervised learning. Temporal Ensembling averages label predictions, while Mean Teacher averages model weights, improving accuracy and reducing the need for labeled data. Mean Teacher offers superior performance on benchmarks, demonstrating the importance of great network architecture in SSL. Another recent work, Virtual Adversarial Training (VAT), proposes a new regularization method based on virtual adversarial loss, a measure of local smoothness of the conditional label distribution given input. VAT represents the adversarial direction without label information, making it appropriate for semi-supervised learning. The computational cost of VAT is relatively low \cite{miyato2018virtualadversarialtrainingregularization}.

SSL methods have demonstrated notable promise in text classification tasks within NLP. For illustration, the application of SSL in scene text detection has led to the development of robust models capable of detecting text with arbitrary forms using small labeled data \cite{liu2020semitext}. Another study exhibited improved text classification accuracy by leveraging SSL methods such as self-ensembling and temporal ensembling, which combine predictions from multiple training epochs to form a prediction \cite{laine2017temporalensemblingsemisupervisedlearning}. Another study applied SSL techniques to enhance the SVM algorithm for text classification, indicating improved classification quality by forming a feature model based on labeled data and improving it with unlabeled data through binary classification \cite{thanh2013text}.

Different SSL methods offer variable levels of effectiveness depending on the application. Graph-based methods surpass relational data handling and have revealed dominance over traditional bag-of-words models in some text categorization tasks \cite{widmann2017graph}. Meanwhile, pseudo-labeling and consistency regularization methods, such as FixMatch, achieve high accuracy by training models with high-confidence predictions on augmented data \cite{sohn2020fixmatch}. Each method has its strengths, and the choice of technique usually relies on the specific requirements and limitations of the task.

One major limitation of SSL is the dependency on high-quality labeled data to generate reliable pseudo-labels. The scalability of SSL methods is a concern, mainly when dealing with large datasets or complex models. Economic considerations also play a role, as obtaining initial labeled data still incurs costs \cite{hady2013semi}.

Arising techniques, such as contrastive learning and adversarial training, present favorable directions for enriching SSL methods \cite{9394423}. Besides, advancements in GNNs and other graph-based approaches are anticipated to improve SSL performance in various applications. Researchers are also exploring better ways to integrate SSL with deep learning frameworks, leveraging both paradigms' strengths to achieve superior results \cite{duarte2023review}.

In this study, we execute a transductive label propagation method based on the manifold assumption to predict labels for an entire dataset in text classification tasks. Our research evaluates the effectiveness of this approach in training deep learning models for sentiment analysis, providing valuable insights into semi-supervised learning. By addressing the boundaries of existing SSL methods and leveraging the manifold assumption, this study aims to enhance the accuracy of text classification models while significantly reducing the costs associated with data labeling.

\section{Fundamentals}
\subsection {Label Propagation with Deep Learning}
Label propagation is a graph-based semi-supervised learning method. In this approach, all data points, whether labeled or unlabeled, are represented as vertices on a graph within a $d$-dimensional feature space. Label propagation considers labeled data as 'sources.' It assigns pseudo-labels to the unlabeled data based on the cluster assumption, which suggests that vertices close to each other on the graph should share similar labels. The ' unlabeled' data becomes useful for further supervised learning by assigning these pseudo-labels to them. This method of assigning pseudo-labels to unlabeled data based on their proximity on the graph facilitates more cost-effective learning \cite{iscen2019label}. 

\begin{algorithm}
\caption{Label Propagation with Semi-Supervised Learning}
\begin{algorithmic}[1]
\REQUIRE Labeled dataset $L$, Unlabeled dataset $U$, Total dataset $T = L \cup U$, Epochs $M$, $E$, $N$
\ENSURE Trained model $F_{\text{Full}}$

\STATE \textbf{Train Baseline Model:}
\STATE $F_0 = \text{TrainBaseline}(L, M)$

\STATE \textbf{Train Fully Supervised Model:}
\STATE $F_{\text{Supervised}} = \text{TrainFullySupervised}(T, M)$

\STATE \textbf{Feature Extraction and Label Propagation:}
\STATE $V = F(F_0, U)$ \COMMENT{Extract hidden representations $V$}
\STATE $\{y_i\}_{i=1}^{l+u} = LP(\{x_i\}_{i=1}^{l+u}, V)$ \COMMENT{Assign or update inferred labels}

\STATE \textbf{Prepare Model for LP-SSL Training:}
\STATE $F_{\text{LP-SSL}} = F_0 - \text{FC}$ \COMMENT{Remove fully connected layer (FC)}

\STATE \textbf{LP-SSL Training:}
\STATE $F_{\text{LP-SSL}} = \text{TrainLP-SSL}(\{x_i, y_i\}_{i=1}^{l+u}, F_{\text{LP-SSL}}, E)$

\STATE \textbf{Complete Training Pipeline:}
\STATE $F_{\text{Full}} = \text{TrainFullPipeline}(L, U, F_{\text{LP-SSL}}, N)$

\RETURN $F_{\text{Full}}$

\end{algorithmic}
\end{algorithm}

The label propagation process, illustrated in (Algorithm 1), involves placing assumed labels $y_i$ to data points $x_i$ within the unlabeled dataset $U$ through the process. This labeling is achieved using the label propagation function $LP$ to the hidden representations $V$, generating $y_i = LP(x_i, V)$. In this procedure, $LP$ utilizes the cluster assumption, considering the closeness of vertices in a graph constructed by data points within a $d$-dimensional feature space. 

Here, $T$ symbolizes the fully labeled dataset, $L$ is the labeled subset, and $U$ is assumed unlabeled. Therefore, all data in $T$ is used for fully supervised training.
In the first stage of the training pipeline, a baseline model is trained on the labeled dataset $L$ for $M$ epochs, as lower-bound, marked as $F_0 = \text{TrainBaseline}(L, M)$. Similarly, a fully supervised model is trained on the labeled dataset $T$ for $M$ epochs, delivering an upper-bound performance indicated as $F_\text{Supervised} = \text{TrainFullySupervised}(T, M)$. Indeed, the baseline model employs a smaller labeled dataset \( L \), but the fully supervised model uses the entire dataset \( T \). The next stage applies feature extraction and process training. At first, the feature extractor $F$ is used on the baseline model $F_0$ and the unlabeled dataset $U$ to get hidden representations $V$, i.e., $V = F(F_0, U)$. Later, label propagation is utilized on $V$ to assign or update presumed labels for unlabeled data points, resulting in $\{y_i\}_{i=1}^{l+u} = LP(\{x_i\}_{i=1}^{l+u}, V)$, where 'l' is the number of labeled points, and 'u' is the number of unlabeled points. 

Further, the model is prepared for label propagation semi-supervised learning (LP-SSL) training by removing the fully connected layer (FC) from the baseline model, stated as $F_\text{LP-SSL} = F_0 - \text{FC}$. LP-SSL then proceeds using both labeled and newly labeled data for $E$ epochs, described as

\begin{equation}
F_\text{LP-SSL} = \text{TrainLP-SSL}(\{x_i, y_i\}_{i=1}^{l+u}, F_\text{LP-SSL}, E)
\end{equation}

Finally, the complete training pipeline includes initializing a model with weights from LP-SSL training and later training it on both the labeled dataset $L$ and the unlabeled dataset $U$ for $N$ epochs, resulting in $F_\text{Full} = \text{TrainFullPipeline}(L, U, F_\text{LP-SSL}, N)$ \cite{iscen2019label}.

In other words, this network is trained with a small set of examples that have labels. Then, using the features learned by the network, it creates a graph where each example is correlated to its closest neighbors. This design helps spread labels from labeled to unlabeled samples in a method known as transductive learning. After setting up this graph, the network is trained again, using all the examples in the dataset. This training includes the originally labeled and unlabeled examples, which now have 'pseudo-labels' assigned. These pseudo-labels are given weights that depend on how confident the network is about the label and how many samples there are of each label, which helps improve the learning process.

\subsection {Dataset}
The dataset utilized in this study is the "Large Movie Review Dataset" (IMDb). It is a leading resource for sentiment analysis and natural language processing tasks, consisting of 50 movie reviews. These reviews are separated into a training set of 25k reviews and a testing set of 25k reviews, balanced between positive and negative sentiments.

Class imbalance, a prevailing issue in text classification tasks, can render label propagation sensitive to imbalances in the labeled data. This sensitivity leads to biased predictions, particularly when certain classes are underrepresented. The balanced nature of the IMDb dataset serves as a safeguard against such issues, guaranteeing an equitable representation of both classes.

\subsection { Data Preprocessing and Cleaning}
For dataset preparation, we present randomness by shuffling the dataset to provide a diverse distribution of data points. The dataset is then separated into training and validation sets with an 80:20 ratio and categorized into labeled and unlabeled subsets to support semi-supervised learning.

During data cleaning, each sentence is converted to lowercase, leading or following spaces are removed, punctuation is separated with spaces, non-letter/punctuation characters are removed, and excessive spaces are cut.

Text data preparation involves three steps: tokenization customized to NLP models, vocabulary building limited to top and unique tokens, and data transformation converting tokenized text into numerical indices aligned with labels for efficient processing.

To enhance model performance, we incorporate pre-trained word embeddings. Word vectors are loaded from a specified file, an embedding matrix is created, and tokens are mapped to token IDs with pre-trained vectors or default vectors for unknown tokens. A vocabulary of the top 10K most frequent tokens is constructed to balance model complexity with linguistic relevance in most configurations.

\subsection{Implementation}
As discussed in the previous section, label propagation approach leverages both the geometrical distribution of the data and the available labels to extend labeling across a dataset. The process starts by calculating the k-nearest neighbors for each data point in the feature space. This involves normalizing the data features and using these to find the closest neighbors, which helps in understanding the local structure of the data. With the neighbors identified, the next step constructs a graph where nodes represent data points and edges are weighted by the distance between these points raised to a power determined by the gamma parameter. This weighted graph is then normalized to balance the influence each node has based on its connectivity. Using the graph, the algorithm propagates labels from the labeled data points to the unlabeled ones. It does this by solving a series of linear equations that diffuse the labels through the graph, smoothing out the label assignments based on the structure captured in the graph. After propagating the labels, each data point is assigned a pseudo label based on the highest probability from the label distribution. The reliability of these pseudo labels is further weighted by their respective entropy, providing a measure of confidence in each label. Finally, to address any class imbalance that might affect learning, the algorithm calculates weights for each class. This is done to ensure that each class is equally represented during any subsequent process.

\section{Experimental Results}
This section outlines the results of our investigations across different processes. We begin by evaluating the impact of various word embeddings on model performance, including Word2Vec, GloVe, and FastText. The analysis highlights the impact of each embedding on key performance metrics. We then examine the outcomes of tokenization variability, specifically focusing on how different token counts affect these metrics. Additionally, we compare the performance of deep learning models—BiGRU, BiLSTM, and 1D CNN. Finally, our exploration of deep learning hyperparameter optimization for the BiGRU model details the effects of settings such as the number of layers, hidden dimensions, and label counts on the accuracy, F1 score, and AUC-ROC.

The outcomes are depicted through comparative bar charts for three baseline and label propagation metrics, with the fully supervised model serving as the upper benchmark. This analysis demonstrates how different embeddings, tokenization strategies, model selections, and hyperparameters influence the resulting performance metrics.

\subsection { Results of Experimenting with Different Word Embeddings}
Using pre-trained embeddings Word2Vec\cite{mikolov2013efficientestimationwordrepresentations}, FastText\cite{bojanowski2017enrichingwordvectorssubword}, and GloVe\cite{pennington-etal-2014-glove} helps to initialize the baseline model with rich semantic information learned from a large corpus of text. This initialization improves the model's performance, especially when dealing with a small amount of labeled data. In the label propagation stage, the model continues from the learned state achieved in the baseline stage, allowing it to benefit from the initially learned embeddings while focusing on improving the labels and model parameters through the label propagation process. 

The experiments use 10 percent of the dataset configured with a BiGRU featuring 64 hidden dimensions and 2 layers, demonstrating the model's adaptability. Label propagation is implemented with 100 nearest neighbors.

\begin{figure}[h]
    \centering
    \includegraphics[width=0.9\linewidth]{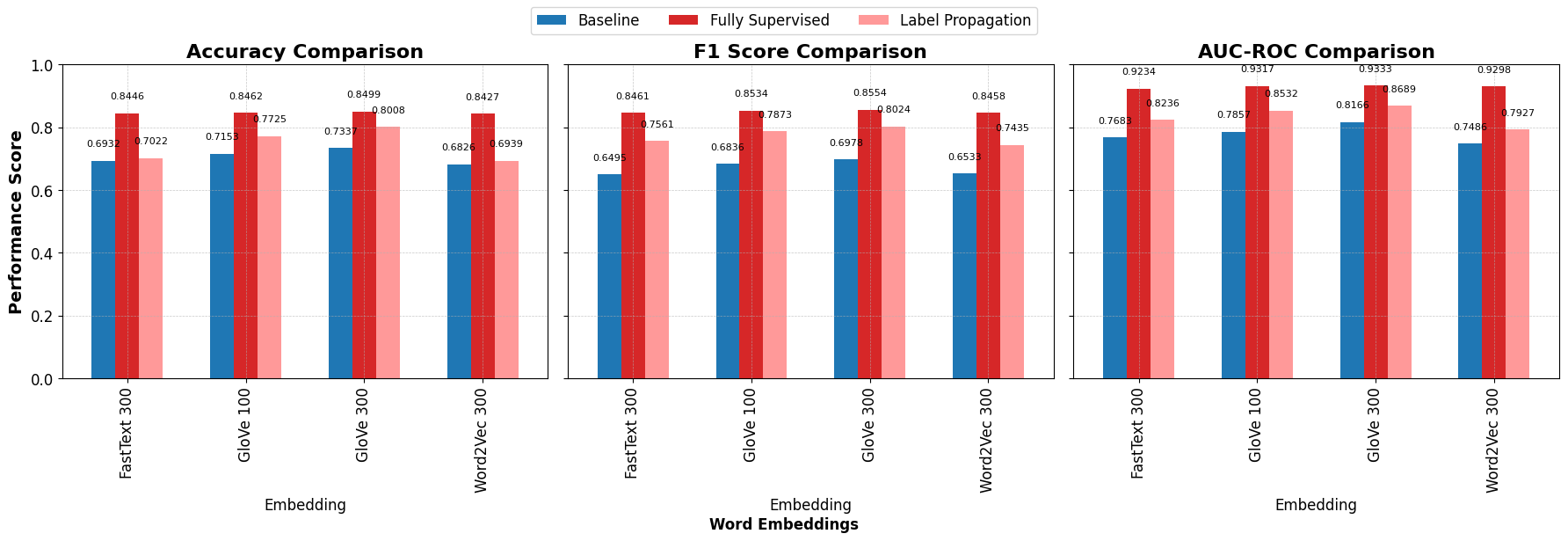}
    \caption{Performance comparison of different word embeddings across Baseline, Fully Supervised, and Label Propagation. The metrics compared are Accuracy, F1 Score, and AUC-ROC. The x-axis represents the type of word embedding used, while the y-axis represents the score for each metric.}
    \label{fig:ssl1}
\end{figure}

In the comparative analysis of four embeddings (Fig. 1), as displayed in the figure, GloVe 300 consistently performs best across all metrics, surpassing FastText 300, GloVe 100, and Word2Vec 300. FastText 300 functions better than Word2Vec 300, but it still requires to keep pace with the results achieved by both GloVe models. GloVe 100 delivers strong results, but GloVe 300 is superior. Word2Vec 300 has the lowest performance overall but is competitive in label propagation. Among the 10,002 tokens analyzed, GloVe accounts for 8,725, FastText for 7,558, and Word2Vec for 7,064. GloVe is the most effective embedding performance, mainly when using higher-dimensional embeddings like GloVe 300. This observation proves that higher-dimensional embeddings enhance model performance by capturing more helpful information.

\subsection { Results of Tokenization Variability on Model Performance}
These experiments, designed to compare the performance with different numbers of tokens (Fig. 2), have shown that 20k tokens consistently outperform 10k tokens across all three metrics (Accuracy, F1 Score, and AUC-ROC) for both the baseline and label propagation models. It is worth mentioning that 12846 out of 20002 tokens are matched with pre-trained fasttext word vectors.

\begin{figure}[h]
    \centering
    \includegraphics[width=0.5\linewidth]{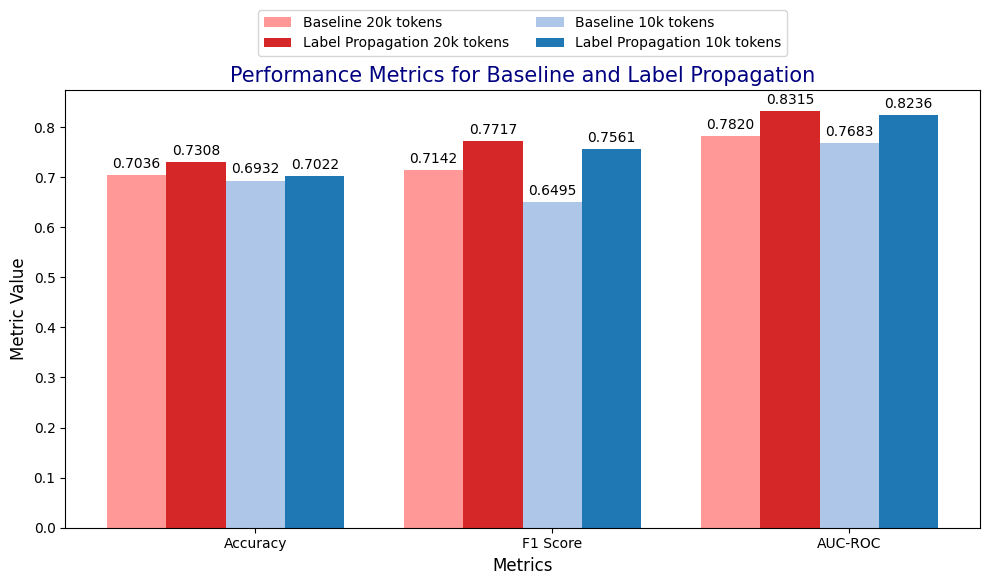}
    \caption{Number of Tokens}
    \label{fig:ssl1}
\end{figure}

\subsection {Comparative Results of Deep Learning Models } 
In this experiment, we employed three distinct models: BiGRU (Bidirectional Gated Recurrent Unit), BiLSTM (Bidirectional Long Short-Term Memory), and 1D CNNs (One-Dimensional Convolutional Neural Networks) to investigate label propagation across all models. (Fig. 3) illustrates model comparisons of BiGRU, 1D CNN, and BiLSTM on 10 percent of data. 

\begin{figure}[h]
    \centering
\includegraphics[width=0.8\linewidth]{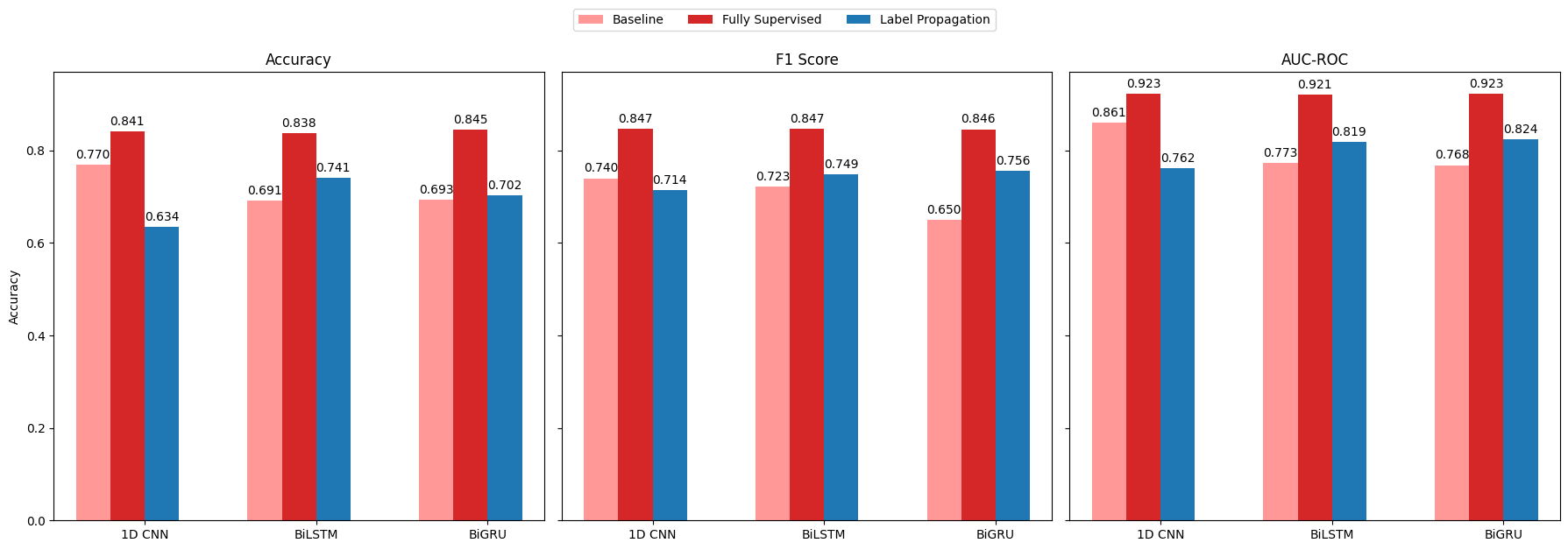}
    \caption{Performance Metrics of 1D CNN, BiLSTM, and BiGRU Models Using All Three Methods}
    \label{fig:ssl1}
\end{figure}

In the baseline model (Fig. 3),  BiLSTM generally performs better regarding F1 Score and AUC-ROC, while BiGRU has a slight advantage in Accuracy. BiLSTM performs better in Accuracy in label propagation, while BiGRU has a slight advantage in F1 Score and AUC-ROC. 1D CNN offers the highest baseline performance across all metrics compared to BiLSTM and BiGRU but performs inadequately in the Label Propagation method. This inadequate performance could be due to the constraints of 1D CNNs in capturing long-term dependencies in text data, which  BiLSTM and  BiGRU models handle better. Indeed, the 1D CNN's performance decreases in the label propagation method due to difficulties in generating accurate pseudo-labels and capturing the global context. This observation stresses the need for models that can effectively handle these intricacies.

\begin{figure}[h]
    \centering
\includegraphics[width=0.7\linewidth]{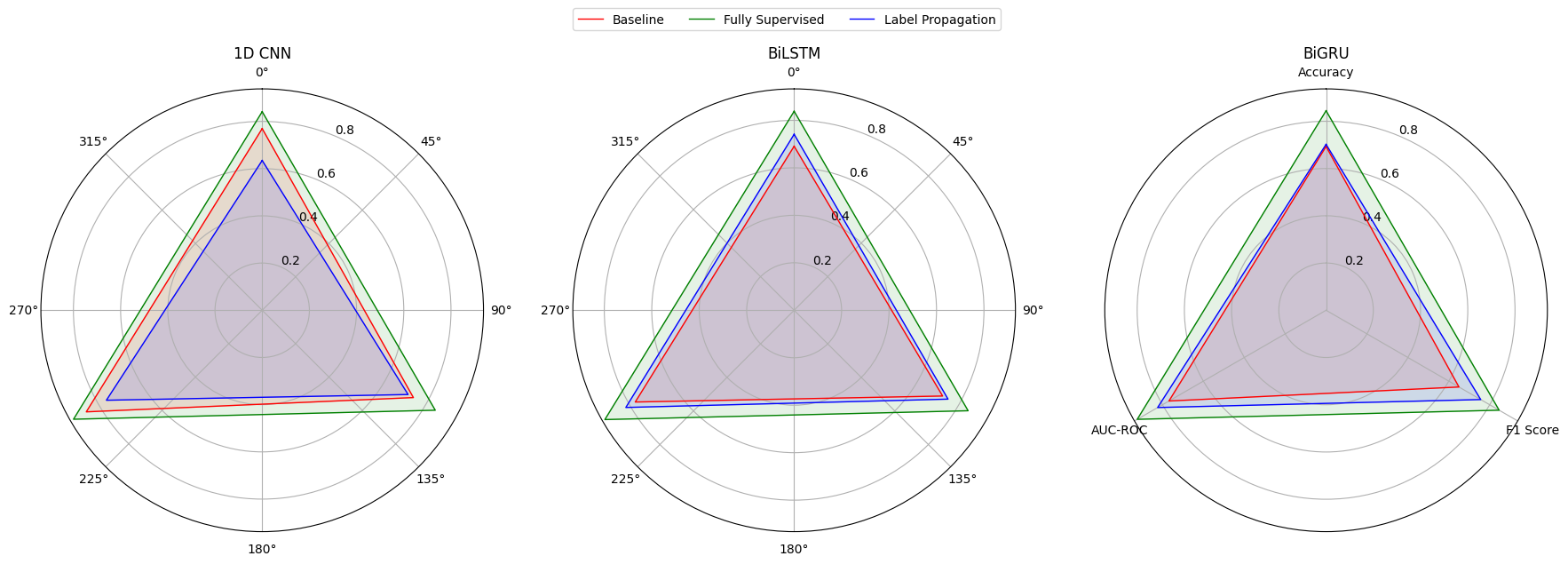}
    \caption{Radar Charts of 1D CNN, BiLSTM, and BiGRU Models for All Three Methods}
    \label{fig:ssl1}
\end{figure}

We have similar observations when comparing Baseline and Label Propagation methods using radar charts (Fig. 4). Both BiLSTM and BiGRU models perform agreeably with Label Propagation, followed by Baseline, highlighting its effectiveness. However, for models like 1D CNN, Label Propagation may only sometimes be beneficial due to its sensitivity to less labeled data and ability to capture local patterns and features.

BiGRU models have been found to perform well due to their ability to capture long-term dependencies and contextual information from both directions in the text.\cite{iscen2019label} BiLSTM models are similar to BiGRUs but can capture even longer dependencies due to the LSTM's memory cell, which also helps mitigate the vanishing gradient problem. BiGRUs and BiLSTMs process the sequence in both forward and backward directions. This process allows the model to understand the context around each word better, considering both past and future information.
 
\subsection {Hyperparameter Optimization Results for BiGRU Model}
This experiment considers the performance of baseline, fully supervised, and label propagation models—across different configurations on a BiGRU model. We examine these methods by varying the neural network layers, hidden dimensions, and data proportions used for label propagation. Although we considered all deep learning metrics, our analysis concentrates on key performance indicators: accuracy, F1 score, and AUC-ROC. The study seeks to determine the effectiveness and efficiency of each variable, delivering insights into their strengths and weaknesses under different experimental setups, mainly focusing on the impact of the number of labels in the label propagation method.

\begin{figure}[h]
    \centering
\includegraphics[width=0.9\linewidth]{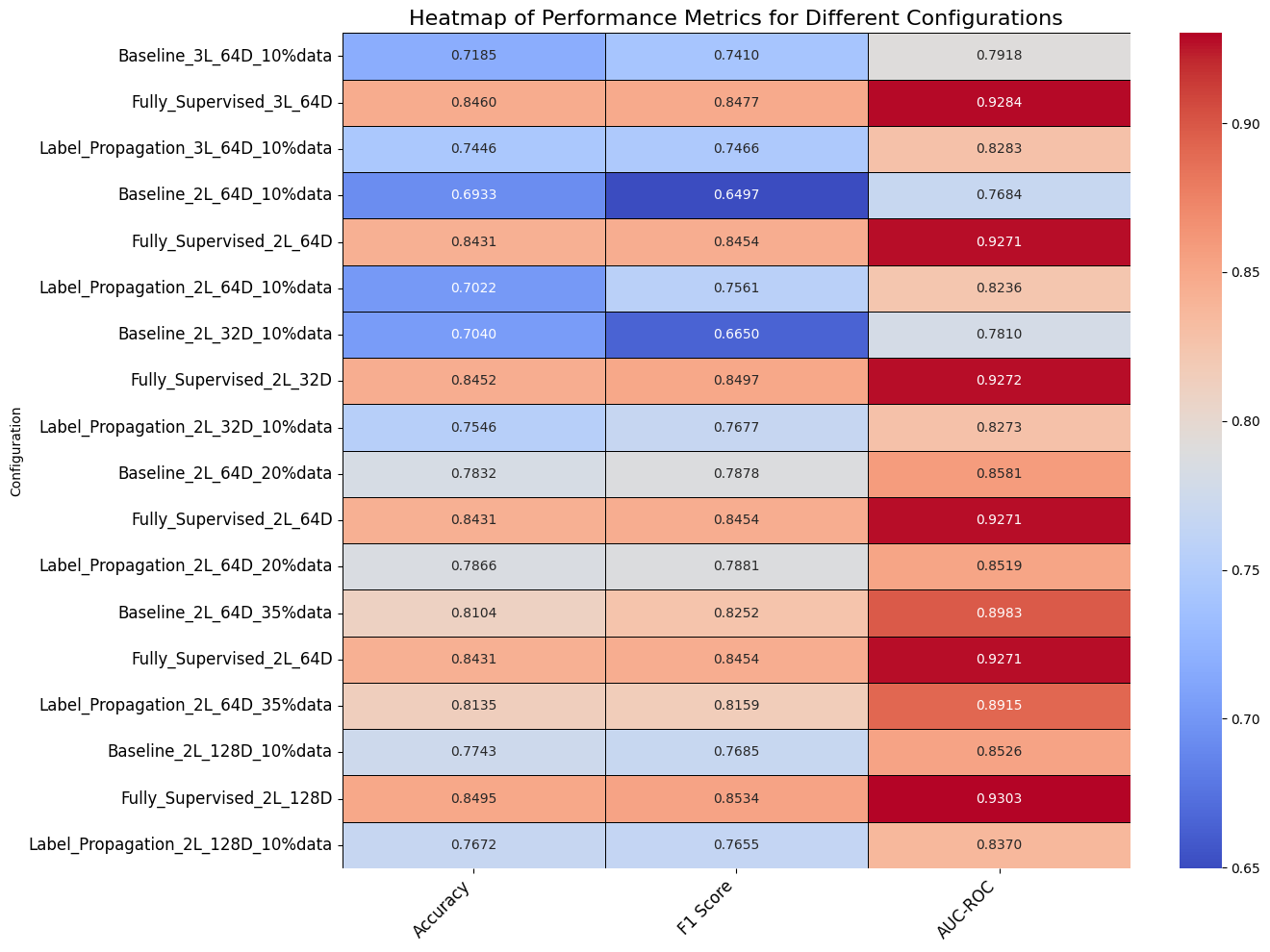}
    \caption{Heatmap of Performance Metrics for Different Configurations on BiGRU}
    \label{fig:ssl1}
\end{figure}

In every tested configuration (Fig. 5), label propagation exceeded the performance of the baseline models, which served as the lower benchmark. Label propagation consistently outperforms the baseline in configurations, showing improvements in F1 Score and AUC-ROC. Its performance is competitive with fully supervised models, which are the upper bound in this comparison. The performance of label propagation models lies between baseline and the fully-supervised models, indicating a robust progression in model performance. The study's success relies on the label propagation-based semi-supervised model, indicating enhancements over the baseline and gaining competitive performance close to the fully supervised benchmark. Expanding the quantity of data (from 10 percent to 20 percent and 35 percent) enhances performance metrics across all methods, with Label Propagation showing notable improvements. Additionally, increasing hidden dimensions (from 32D to 128D) positively impacts performance as well, particularly for Fully Supervised and Label Propagation methods, highlighting the benefits of richer feature representations.

\begin{figure}[h]
    \centering
\includegraphics[width=0.6\linewidth]{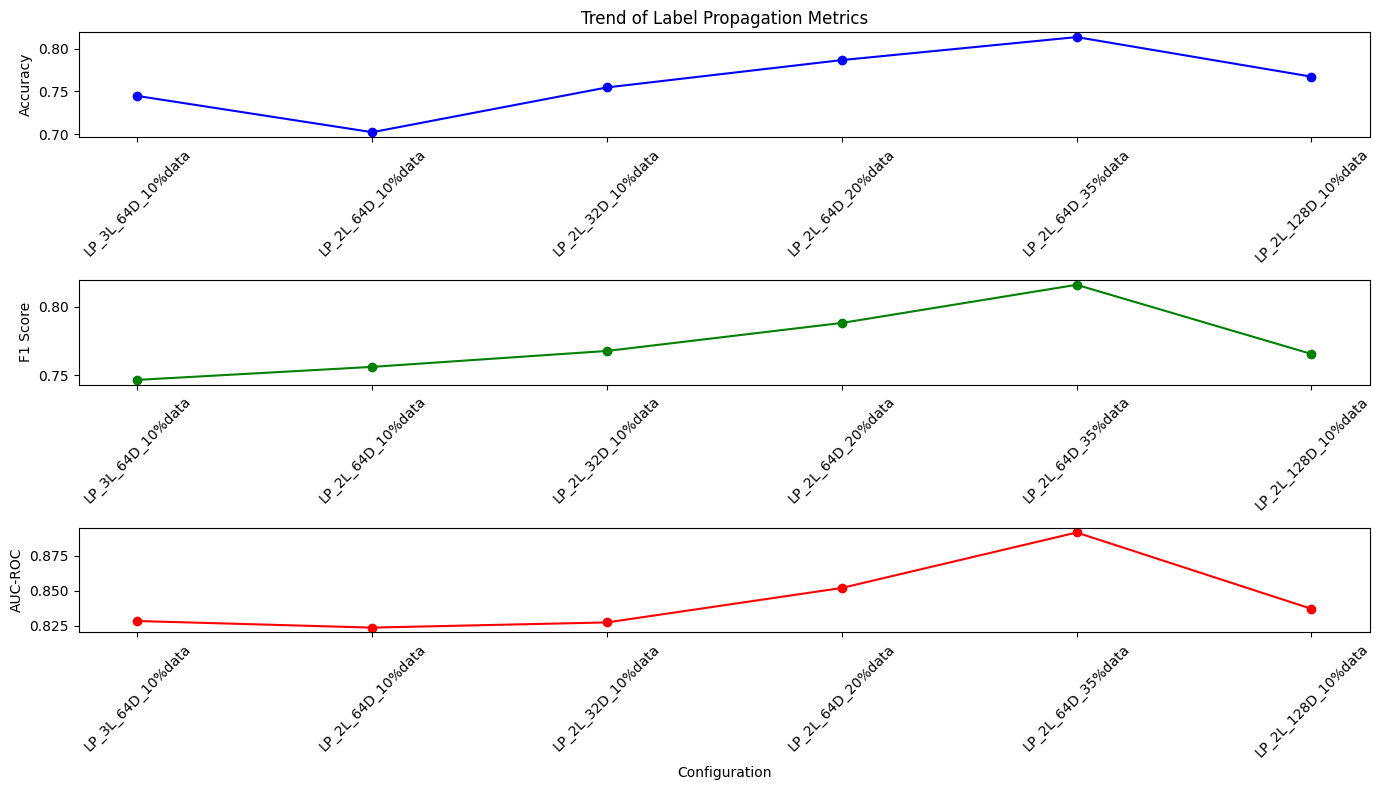}
    \caption{Performance Metrics Trend for Label Propagation Configurations}
    \label{fig:ssl1}
\end{figure}

All metrics follow a similar trend (Fig. 6) of initial improvement with a rising data ratio and hidden dimensions, followed by a performance decrease in the last configuration.

\begin{figure}[h]
    \centering
\includegraphics[width=0.5\linewidth]{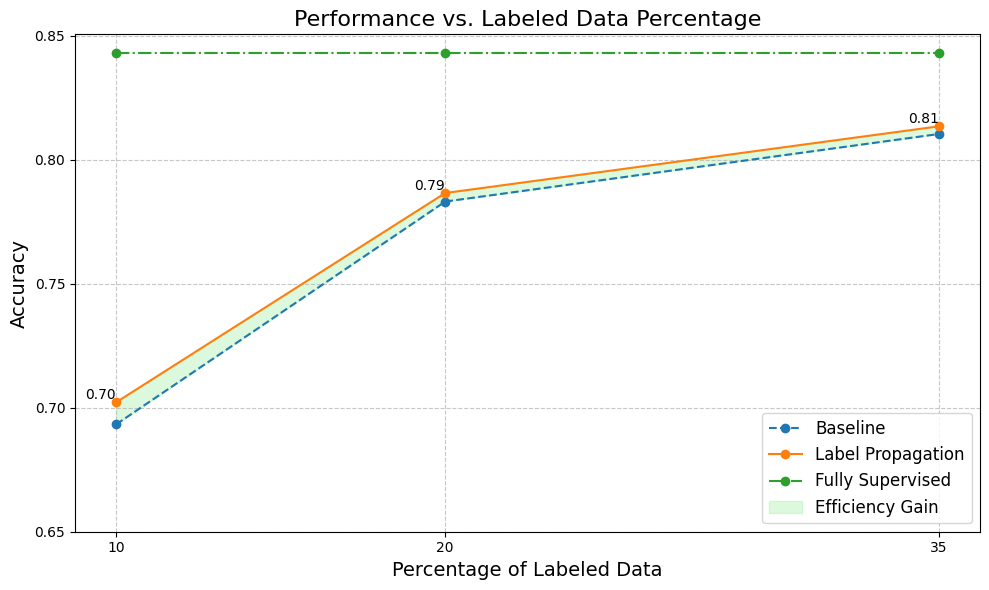}
    \caption{Accuracy vs. Labeled Data Percentage for Models, Showing Efficiency Gains with Label Propagation}
    \label{fig:ssl1}
\end{figure}

Across all three metrics (Fig. 7), increasing the proportion of labeled data (from 10 percent to 35 percent) generally improves the performance of Label Propagation methods. However, the critical fact is that the efficiency gain of Label Propagation over the Baseline diminishes as the amount of labeled data increases. 

The quantity of labeled examples used in Label Propagation is pivotal. While increasing the number of labeled examples generally enhances performance, an excessive amount reduces the model's effectiveness. This occurs because the model begins to overfit to the labeled data, diminishing the benefits of semi-supervised learning, a powerful approach that relies on leveraging both labeled and unlabeled data for more reasonable generalization.

\subsection{Discussion and Limitations}
Label propagation encounters several challenges in text classification tasks that require attention. A primary issue is sensitivity to noise, as text data often possesses inaccuracies from misspellings, abbreviations, or ambiguous language. This issue is especially acute with user-generated or unstructured text. 

Additionally, label propagation needs more discriminative capability when applied to text data with complex, non-linear relationships. This problem is also evident in tasks involving complicated language, where defining precise decision boundaries for accurate classification is challenging.

Another noteworthy problem for label propagation in text data is the reliance on an efficacious graph structure. Creating a graph representing relationships between text instances affects the propagation process's quality. Likewise, the computational complexity of label propagation poses challenges, especially with the large datasets typical in NLP tasks. Forming and updating graphs can be computationally intensive for large text corpora.

Hyperparameter tuning is vital in label propagation for text classification tasks. It involves optimizing parameters specific to constructing text-based graphs and the propagation processes. The choice of hyperparameters, such as the similarity measure and the number of neighbors in the graph, dramatically influences the model's performance, making the tuning process complex yet required for achieving optimal results in sentiment analysis. Determining the optimal number of k nearest neighbors is essential for achieving accurate results, which can be done similarly to the approach employed in \cite{jafarlou2022improvingfuzzylogicbasedmapmatching}. Indeed, too few neighbors may introduce noise through inaccurate labels, while too many could result in excessively smoothed labels, affecting the overall model performance. Additionally, hyperparameters, such as the number of hidden dimensions, can cause performance to plateau or even decline if not paired with adequate data.

Addressing these complex challenges will enhance the effectiveness and broaden the applicability of label propagation in text classification.

\section{Conclusion}
In conclusion, using unlabeled data alongside label propagation techniques has significantly enhanced semantic analysis performance, specifically when the number of labeled data was few. This research comprehensively evaluates label propagation's role, effectiveness, and practicality in enhancing sentiment analysis text classification tasks. The performance evaluation compares the model's performance against baseline and fully supervised models. The project's success is evident in the semi-supervised model based on label propagation, which improved over the baseline. Importantly, our ongoing project is committed to refining label propagation algorithms by integrating extra data, optimizing graph construction for better performance, incorporating domain adaptation strategies, and implementing regularization techniques to enhance robustness. In future work, we plan to combine label propagation with active learning strategies to identify and label the most insightful unlabeled data points to maximize performance yields while minimizing labeling endeavors.

\bibliographystyle{ACM-Reference-Format}
\bibliography{sample-acmsmall-submission.bib}

\end{document}